\documentclass[10pt,twocolumn,letterpaper]{article}

\usepackage{times}
\usepackage{epsfig}
\usepackage{graphicx}
\usepackage{amsmath}
\usepackage{amssymb}

\usepackage{subfigure}

\usepackage{algorithm}
\usepackage{algorithmic}
\usepackage{hyperref}

\newcommand {\R}{\mathbb {R}}

\newcommand {\argmax}{\mathrm{argmax}}
\newcommand {\argmin}{\mathrm{argmin}}

\begin{document}

%%%%%%%%% TITLE
\title{Fast approximations to structured sparse coding and applications to object classification}

\author{Arthur Szlam\\
\and
Karol Gregor\\
\and
Yann LeCun\\
}

\maketitle
% \thispagestyle{empty}
%%%%%%%%% ABSTRACT
\begin{abstract}
We describe a method for fast approximation of sparse coding.   The input space is subdivided by a binary decision tree, and we simultaneously learn a dictionary and assignment of allowed dictionary elements for each leaf of the tree.  We store a lookup table with the assignments and the pseudoinverses for each node, allowing for very fast inference.
%During inference 
%the algorithm uses a tree to assign an input to a group of allowed dictionary elements and then finds the corresponding coefficient values 
%using a cached pseudoinverse. We give an algorithm for learning the tree, the dictionary and the dictionary element assignment, and 
In the process of describing this algorithm,  we discuss the more 
general problem of learning the groups in group structured sparse modelling.
We show that our method creates good sparse representations by using it in the object recognition framework of \cite{lazebnik06,yang-cvpr-09}. Implementing our own fast version of the 
SIFT descriptor the whole system runs at 
$20$ frames per second on $321 \times 481$ sized images on a laptop with a quad-core cpu, while sacrificing very little accuracy on the Caltech 101 and 15 scenes benchmarks.
\end{abstract}

%%%%%%%%% BODY TEXT
\section{Introduction}
\label{Intro}
Sparse modeling \cite{olshausen1996emergence,elad-ksvd} has proven to be a useful framework for signal processing.  Each point 
%$x$ 
from a dataset consisting of vectors in a Euclidean space
%$X\subset \R^d$ 
is represented by a vector 
%$z$  
with only a few nonzero coefficients.  Sparse modeling has lead to state of the art algorithms in image denoising, inpainting, supervised learning, 
and of particular interest here, object recognition.   
The systems described in \cite{lazebnik06,yang-cvpr-09,koray-psd-08,mixture_model_eccv_2010,ylan-iccv-2011} use sparse coding as an integral element. Since the coding is done densely in an image with
 relatively large dictionaries, this is a computationally expensive part of the recognition system, and a barrier to real time application. 
%repeat once [Stanford] or twice [Y-Lan, Kai Yu] the following set of operations: sparse feature extraction at every location of the input followed by pooling of each feature over a small neighborhood. Essentially linear classification is then used to obtain labels. Sparse feature extraction is the computationally expensive element of these systems.
%because 1) input typically needs to be compared to each feature vector and 2) there are many possible sets of nonzero coefficients that can represent the input. 
The main contribution of this paper is a fast approximate algorithm for finding sparse representations; we use this
 algorithm to build a system with near state of the art recognition performance that runs in real time.   During inference 
the algorithm uses a tree to assign an input to a group of allowed dictionary elements and then finds the corresponding coefficient values 
using a cached pseudoinverse. We give an algorithm for learning the tree, the dictionary and the dictionary element assignment, and along the way discuss methods for the more 
general problem of learning the groups in group structured sparse modelling.

One standard formulation of sparse coding is to consider $N$ $d$-dimensional real vectors $X=\{x_1,\ldots,x_N\}$ and represent them using $N$ $K$-dimensional real vectors $Z=\{z_1,\ldots,z_N\}$ using a $k \times d$ dictionary matrix $W$ by solving
\begin{equation}
\argmin_{Z,W} \sum_k||Wz_k-x_k||^2, \,\, \text{s.t.}\,\, ||z_k||_0\leq q,\label{l0}
\end{equation}
where $||\cdot||_0$ measures the number of nonzero elements of a vector;  each input vector $x$ is thus represented 
as a vector $z$ with at most $q$ nonzero coefficients. While this problem is not convex, and in fact the problem in the $Z$ variable is 
NP-hard, there exist algorithms for solving both the problem in $Z$ (e.g. Orthogonal Matching Pursuit, OMP \cite{}) and the problem in 
both variables (e.g. $K$-SVD \cite{elad-ksvd}) that work well in many practical situations.

It is sometimes appropriate to enforce more structure on $Z$ than just sparsity.   For example, many authors have noted that the solution to the $z$ minimization in 
\eqref{l0}  (and its $l_1$ relaxation) is very unstable in the sense that nearby inputs can have very different coefficients, in part because of the combinatorially 
large number of possible active sets (i.e. sets of nonzero coordinates of $z$).  
This can be a problem in classification tasks.  Other times we may know in advance some structure in the data that the coefficients should preserve.   
Various forms of structured sparsity are explored in \cite{jenattonProxHierarchical2010,kim_xing_group_lasso,Jacob:group,model_based_cs_arxiv}.

A simple form of structured sparsity is given by specifying a list of $L$ allowable active sets, and some function $g:\R^d\mapsto \{1,...,L\}$ associating to each $x$  
to one of the $L$ configurations.  An  example of this is the output of many subspace clustering algorithms.  There, $X$ is reordered and partitioned into $PX=[X_1\,\, X_2 ... \,\,X_L]$ 
(where $P$ is a permutation matrix), so that each block $X_j$ is near a low dimensional subspace spanned by $B_j$.  
 Supposing for simplicity that each of the $B_j$ are of the same dimension $q$, then if we set $W=[B_1 ... B_L]$, 
the allowable active sets are given by $\{1,...,q\}$, $\{q+1,...,2q\}$, etc.   
By setting the allowable active sets to the blocks, and the function $g$ to simply map each point to its 
nearest subspace (say in the standard sense of Euclidean projections), then we get an example of structured sparsity as described above; this sort of method is used 
in object recognition in \cite{mixture_model_eccv_2010}.

%Our algorithm is related to group sparsity [...]. The basic goal of group sparse coding 
%is to implement the idea that some dictionary elements are more closely related then others. For example if the input consists of images of 
%handwritten digits $0,\ldots,9$ it is useful to group some of the coefficients into ten groups, one group for each digit [...]. When an input is 
%presented, the system will try to represent it with dictionary elements coming mostly form one of the groups. This can help generalization 
%for example for de-noising or recognition. In our case we force the input to be represented strictly by one group. 
%Mathematically, we have a set of $n_g$ groups of indices of $z$. Let $g_{\alpha,i}$ denote the index for groups 
%$\alpha \in \{1,\ldots,n_g\}$ and $i \in \{1,\ldots, n_z\}$ where $n_z$ is the number of elements in a group.  A given 
%input $x$ is assigned one $\alpha$ and then represented by the elements $z_{g_{\alpha,i}}$, $i=1,\ldots,n_z$. Our first 
%contribution is an algorithm that at the same time assigns which dictionary elements should be in which group and learns 
%the dictionary elements ideal for this assignment and vice versa.

In this work we will try to learn the $L$ configurations as well as the dictionary.  We introduce a LLoyd-like algorithm that alternates between updating the dictionary, updating the assignments of each data point to the groups, and updating the dictionary elements associated to a group 
via simultaneous OMP \cite{gisttr05}. 

At inference time, we need a fast method for determining which group an  $x$ belongs to. 
This is computationally expensive if there is a large number of groups and one needs check the projection onto each group.  However, by specializing the Lloyd type algorithm to the case when each group is composed of a union of (perhaps only one) leaves of a binary decision tree, we will build a fast inference scheme into the learned  dictionary.
The key idea is that by using SOMP, we can learn which leaves should use which dictionary elements as we train the dictionary.  To code an input, we march it down the 
tree until we arrive at the appropriate leaf. In addition to the decision vectors and thresholds, we will store a lookup table with the active set of each leaf as learned above, and the pseudoinverse of the columns of $W$ corresponding to that active set.   Thus after following $x$ down the tree we need only make one matrix multiplication to get the coefficients.   

%Then we minimize a least 
%square problem using the columns of $W$ given by the  is precomputed resulting in one matrix multiplication.

%In fact the original formulation
% eq. (\ref{l0}) is an example: groups are formed by all possible combinations of $q$ indices (although that is not really a group sparsity). 
%Essentially one builds a tree for deciding which group to go to. The first level of the tree splits all of groups in two halves. 
%The second level splits each of the two halves further and so on. To find a group for a given input, we have a decision boundary at every level 
%of the tree. The leaves of the tree are the final groups each with a set of dictionary indices. The decision boundaries are learned by k-means.

Finally, we would like use these algorithms to build an accurate real time recognition system.  We focus on a particular architecture studied in \cite{lazebnik06,yang-cvpr-09,mixture_model_eccv_2010,ylan-iccv-2011}. First, SIFT descriptors are calculated densely over the image. Then (a form of) sparse coding is used to calculate a sparse vector at every location from the 
corresponding sift vector. Then each feature is pooled over a small number of spatial regions and the results are concatenated. 
Finally the labels are obtained using linear SVM or logistic regression. 

We use this pipeline with two modifications. First we write our own fast implementation of the SIFT descriptor. Second 
we use our fast algorithm for the sparse coding step. The resulting system achieves nearly the same performance as 
exact sparse coding calculation but processes $321 \times 481$ size images at the rate of $20$ frames per second on a laptop computer with a quad core cpu.

The rest of this paper is organized as follows:  Section 2, we discuss greedy structured sparse modeling, and describe in depth how to train a model that 
learns the structure, and that respects a given set of groups given by a tree.
In section 3, we show experiments on image patches to qualitatively demonstrate what learned groups look like and then we apply our methods to object recognition.

\section{Hashing and dictionary learning}
\subsection{A simple form of structured dictionary learning}
\label{sec:sgs}
Here we will first suppose that a list of $L$ perhaps overlapping groups $G_1, ... ,G_L$ on the coefficients $Z$ is given.   That is, if we are learning a representation of $X$ with $K$ atoms, each $G_i\subset\mathcal{P}(\{1,...,K\}$, where $\mathcal{P}$ is the set of all subsets of its argument, is specified.   We can generalize the LLoyd algorithm for $K$ means or $K$ flats to this setting.  After initializing the dictionary $W$, we find the distance of each $x$ in $X$ to its projection $P_{G_i}x$ onto the span of $W_{G_i}$ for each $i$.  Each $x$ is associated to the $i$ with the smallest distance
\begin{equation} x\mapsto \argmin_{i\in \{1,...,L\}} ||P_{G_i}x-x||^2, \label{group_choice}\end{equation}
 and we find the coefficients \[z%=Ax
=(W_{G_i}^TW_{G_i})^{-1}W_{G_i}^Tx.\]  
Then we update $W$ to be the minimum of the convex problem \[\argmin_{W}
%%{||W_j||^2\leq 1 \forall \text{  columns $j$}} 
\sum_x||Wz-x||^2,\] and repeat.  Each of the subproblems either has an explicit solution or is convex, and so the energy decreases.  When the training is finished, we define $g$ to be the function that maps each point in $x\in \R^d$ to the $i$ minimizing the error of the projection of $x$ onto the span of $W_{G_i}$. 

We can also run the same sort of algorithm when in addition to each group $G$ specifying a list of indices, it also specifies a cost for the use of the dictionary elements associated to each index.  If we choose an $l_2$ cost for each of the coefficients, we still get explicit updates and the decrease of energy at each round.

Note that if the number of groups is very large, it may be too costly to find the best group for each $x$ exaustively.   However, we can make a greedy approximation by running a modified OMP.  Here, supposing at iteration $s$ of the OMP we have an active set $\Omega$, the available dictionary elements to add to $\Omega$ are the union of all groups containing $\Omega$.   It is not necessary to be able to enumerate all the groups to use this method, only to have a subroutine which given $\Omega\subset\{1,...,k\}$ can return $\bigcup_{\Omega \subset G} G$.   However, using this sort of greedy approximation removes the guarantee that the energy decreases at each iteration.

\subsection{Learning the groups with simultaneous orthogonal matching pursuit}
\begin{figure}
\begin{minipage}[t]{0.45\textwidth}
\begin{algorithm}[H]%this H is important
   \caption{SOMP \cite{gisttr05}}
   \label{alg:somp}
\begin{algorithmic}
    \FUNCTION{${Z=\rm\bf SOMP}(X,W,K)$}
   % \STATE {\bf Require:} $k>0$ number of coefficients.
    \STATE {\bfseries Initialize:} coefficients $Z=0$, residual $R=X$, active set $\Omega=\emptyset$.
    \REPEAT
     \STATE $j=\argmax_i \sum_s|W_i^TR_s|$
     \STATE $\Omega=\Omega \bigcup j$
     \STATE $Z=\left(W_{\Omega}^TW_{\Omega}\right)^{-1}W_{\Omega}^TX$
     \STATE $R = X-WZ$
    \UNTIL{$K$ iterations}
    \ENDFUNCTION
\end{algorithmic}
\end{algorithm}
\end{minipage}
\end{figure}
In the previous section the groups were specified in advance.  If we want to learn the groups, we can add a step in the algorithm.   Now instead of taking the list of groups as input, we instead input just the number $K$ of dictionary elements and the number of coefficients allowed per $x$.  After associating to each $x$ the group that best represents it, we can turn around and consider all the $x$ associated to that group.  Our task is then to choose a subset of the dictionary that best represents that group.  A greedy approximation to this problem in the least squares sense is given by the Simultaneous Orthogonal Matching Pursuit algorithm (SOMP)\cite{gisttr05}.  This algorithm proceeds just as a standard OMP, but at each iteration, all the $x$ associated to a given group have to choose the next dictionary element added to the group together.  See algorithm \ref{alg:somp}.  

Unfortunately, because neither OMP nor SOMP is guaranteed to find the optimal solution to the NP hard problems they address, the energy may not decrease at each iteration with this scheme; however, as usual, we have found that in practice these methods do usually lead to a decrease in the energy.   As in $K$-means, it may happen that no group uses a dictionary element; in such a situation one can remove a dictionary element from one of the groups, find the residual, and replace the unused dictionary element by the principal component of the residual.

We note that the model presented here can be thought of as a greedy sparse coding version of a ``topic model''.  The dictionary elements act as the words, the $x$ as the documents, and the groups are the topics.   The algorithm learns the topics and the dictionary simultaneously. 
%discuss shared parts and pooling?

\subsection{Hashing, quantization, and dictionary learning}

\begin{figure}
\begin{minipage}[t]{0.45\textwidth}
\begin{algorithm}[H]%this H is important
   \caption{Learning a dictionary and groups }
   \label{alg:h_d}
\begin{algorithmic}
    \STATE {\bf Require:}  data $X$, number of dictionary elements $K$, number of active coefficients per data point $q$, number of iterations $I$, and if desired, $g:\R^d\mapsto \{1,...,M\}$.
    %\STATE {\bfseries Initialize:} coefficients $Z=0$, residual $R=X$, active set $\Omega=\emptyset$.
    \REPEAT
     \STATE  1: Each $x$ chooses a group in $\{1,...,L\}$ via \eqref{group_choice}  or by the modified OMP as in Section \ref{sec:sgs}.  If $g$ is given, all the $x$ in a hash bucket are forced to choose the same group.
    \STATE 2: Each group in $\{1,...,L\}$ chooses subset of $\{1,...,K\}$ using ${Z=\rm\bf SOMP}(X,W,k)$. 
     \STATE 3: Update $W$, either via $K$-SVD, or a least squares solve.     
     \UNTIL{$I$ iterations}
 %   \ENDFUNCTION
\end{algorithmic}
\end{algorithm}
\end{minipage}
\end{figure}

The main focus of this work will be choosing a $g$ that can be computed rapidly and learning a dictionary that respects $g$.  We will consider $g$ to be a hash function on $\R^d$, and hash buckets will be the atomic units of the groups; that is, the groups will either be the hash buckets or will be glued together from the hash buckets.  This can be considered a sort of geometric regularization of the sparse coding problem: the active set will be forced to remain constant on the region of $\R^d$ corresponding to each hash bucket. 

Once $g$ is chosen, we will learn the dictionary (and perhaps groups) as above, but instead of allowing each $x$ to choose the group that best represents it individually, the $x$ in a hash bucket will need to choose the group that best represents them together on average.  We will also try to approximate standard greedy dictionary learning; in this case, there will be one group for every hash bucket.  As above, and as with $K$-means, it may happen that no spatial bucket uses a particular group; in that case we can just pick a bucket at random and use the output of SOMP on that bucket to regenerate the unused group.

Learning how to quantize $\R^d$ is a much studied (but still not completely understood) problem.  One common motivation is to build a data structure allowing nearest neighbors from a given data set to be quickly computed.   Another common motivation is to use the buckets of the quantization as words to build bag of words feature representations.   
The relationship between vector quantization and sparse coding has studied before by many authors\cite{}.  In particular, $K$-means is simply $l_0$ sparse coding with only the coefficients $0$ and $1$ allowed, and only 1 nonzero per $x$\footnote{``shape gain coding'' allows a non-binary coefficient}.

In this work we will use a $2$-means tree \footnote{Although perhaps not exactly standard usage, we will call the data structure obtained from binary partitions of $\R^d$ a hash} with subdivisions along medians to define $g$.   We start by taking the entire data set and running $2$-means, obtaining centers $c_1$ and $c_2$.  We take each data point $x\in X$and find the angle between $x$ and $c_1-c_2$; $X$ is divided at the median.  We then repeat on each of the pieces, continuing until each piece is within a given distance to its mean, or a set depth $p$, whichever comes first.  
We initialize the $2$-means with farthest insertion, as in \cite{orss}.  Note that our experience is that very few iterations are necessary, and really the farthest insertion is suffient; in fact cutting in random directions (with some additional tricks and randomizations) has been shown to lead to good partitions when the underlying data has a ``manifold'' structure, see \cite{tree_hash}.  The number of buckets at the bottom of the tree is upper bounded by $2^p$; we will choose $p$ small enough so that it is simple to store a lookup table with the indices into the dictionary for each bucket, as well as the decision vectors for each branch in the tree.

We also could use mappings of the form $g(x)=s(h(Hx+b))$, where $H$ is a $p\times d$ matrix, $h$ is some sort of nonlinearity (e.g. $\tanh$, or $\sin$), $b$ is an offset, and $s$ is a thresholding function \cite{}.  These mappings require less storage and are somewhat simpler to compute for the same bit depth, but on the data sets we work on, they have the disadvantage that many of the buckets are often empty or have very few entries for reasonable $p$.  While this can be remedied by simply gluing (nearly) empty buckets to nearby full buckets and updating the lookup table, we have found the trees to work better.  Note also that unlike in nearest neighbor data structures, it is unnecessary for leaf nodes to keep track of spatially nearby leaves that are far away in the tree metric, because all we care about is which dictionary atoms are used at that node.

After building $g$ and training the dictionary, in order compute the coefficients of a new data point $x$, we pass it though the tree, obtaining $g(x)$.  We lookup $g(x)$ in a table, and this gives an index of $m$ columns $\Omega$ of $W$; at this point we solve the linear system $W_{\Omega}z-x$ to get the outputs.  Alternatively, for each group, we can store $(W_{\Omega}^TW_{\Omega})^{-1}$ (or some stable factorization), or $(W_{\Omega}^TW_{\Omega})^{-1}W_{\Omega}^T$, and just do the requisite matrix multiplications
\subsection{Discussion of related work} 
%There has been an explosion of work on sparse coding and dictionary design, and in particular in object recognition in images this has been recognized as a very useful technique.  

The idea of clustering the input space and then using a different dictionary for each cluster has appeared several times before.  As mentioned in the introduction, a simple example is the $K$-flats algorithm, or other subspace clustering algorithms \cite{vidal_subspace}.  There, the subdictionaries serve the dual purpose of determining the clusters and also finding the coefficients for the data points associated to them.   More recently this technique has been succesfully applied to object recognition by \cite{mixture_model_eccv_2010,ylan-iccv-2011}.  In those works, the clusters are determined by $K$-means (or a Gaussian mixture model); in the first, there is a different dictionary for each cluster, and the code is the size of the union of all the subdictionaries, but only the blocks corresponding to the centroids near the input are nonzero.  In the second work, the dictionaries for each centroid are the same, but the code is still a concatenation of the codes associated to each centroid (and are set to zero if the input does not belong to that centroid).  The current work differes from these in two ways.  The first is the use of a fast method for clustering, and the second is the use of shared parts across the dictionaries, where the organization of the parts sharing has been learned from the data.

In \cite{geometric_wavelets} the authors construct a dictionary on the backbone of a hierarchical clustering with fast evaluation.  They also use shared parts.  However, in that work the part sharing is determined by the tree structure of the clustering, and not learned.

There is now a large literature on structured sparsity.
Like this work, \cite{model_based_cs_arxiv,huang_zhang_metaxas_structure} use a greedy approach for structured sparse coding based on OMP or CoSaMP.
%look for v cevher other works...
Unlike this work, they have provable recovery properties when the true coefficients respect the structure, and when the dictionaries satisify certain incoherence properites.  On the other hand, those works do not attempt to learn the dictionary, and only discuss the forward problem of finding $z$ from $x$ and $W$.The works in \cite{jenattonProxHierarchical2010,kim_xing_group_lasso,Jacob:group} use an approach to structured sparsity that allows for convex optimization in $z$.  In these works the coefficients are arranged into a predetermined set of groups, and the sparsity term penalizes the number of active groups, rather than the number of active elements; the dictionary is trained to fit the data.   None of these works attempt to learn the group structure along with the dictionary

Finally we note that other works have explored the idea of accelerating sparse coding by training the dictionary along with an approximation method, e.g. \cite{koray-psd-08,lista}.  
In the first, the approximation is via a single layer feed forward network, and in the second, via a multilayer feed forward network with a shrinkage nonlinearity.  
This work uses a tree and lookup table instead.

\section{Experiments}

\subsection{What do the groups look like?}
To get a sense of what kind of groups learned from algorithm \ref{alg:h_d} look like, we train a dictionary on 500,000 $8\times 8$ image patches, and view the results.  The image patches are drawn from the PASCAL dataset, and their means are removed.  We train a dictionary with 256 elements and 512 groups; each group has 5 dictionary elements in it.   We train using the batch method, with a $K$-SVD update for the dictionary.

\begin{figure*}
\begin{center}
\includegraphics[width=.9\textwidth]{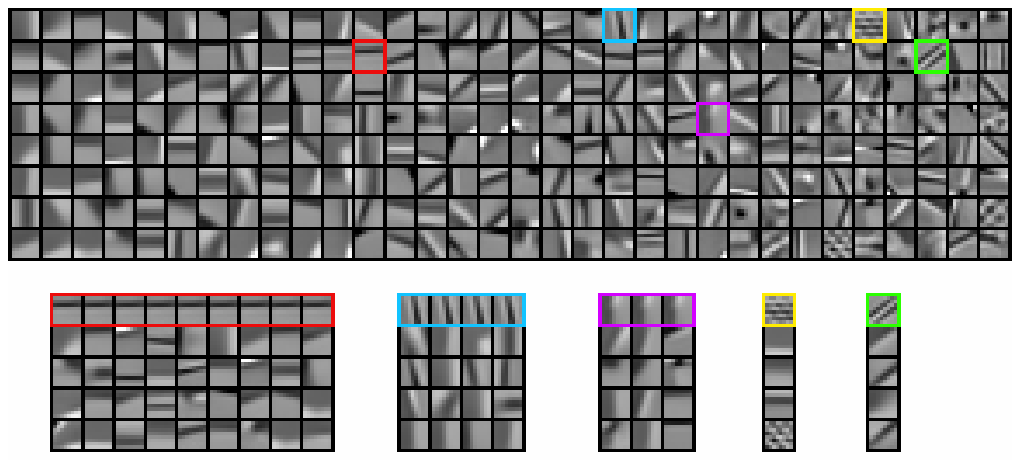}
\end{center}
\caption{256 dictionary atoms in 512 groups trained by algorithm \ref{alg:h_d} on 500,000 $8\times 8$ image patches.  The group structure and dictionary were trained simultaneously.  The dictionary elements, shown on top, are ordered by popularity (the number of groups they belong to).  Underneath, for each dictionary atom in a colored square, we show all of the groups containing it.  These groups can be thought of as ``topics''.  Less popular atoms tend to be more specialized.}
\label{fig:p_ordered}
\end{figure*}

After training, some of the dictionary elements are used by many groups, and others are used by only a few.   The median number of groups using a given element is 6; 47 elements are in exactly 1 group, and 15 are in more than 30.  In figure \ref{fig:p_ordered} we display the dictionary ordered by the number of groups containing each element; this number increases in each column and moving to the right.  Unsurprisingly, ``popular'' elements that belong to many groups are low frequency.  In this figure we also show the groups containing a few chosen atoms.

\subsection{Review of the image classification pipeline}

Here we will review a standard pipeline for object recognition \cite{lazebnik06,yang-cvpr-09}, while giving details about our implementation, which streamlines certain components.
It consists of the following parts: 1) Calculation of sift vectors at every location (sift grid) 2) Calculation of the feature vectors for every sift vector using the "tree sparse coding" described above, 3) Spatial pyramidal max pooling 4) logistic regression or SVM classification.  Care is taken to calculate each of these parts efficiently.  

\subsubsection {Sift grid}

We run tests with two different implementations of dense sift.   The first is matlab code by L. Lazebnik \cite{lazebnik06}.   We also use a fast, approximate c++ version that we coded ourselves.  The details are as follows:
%The following set of operations calculates sift vector at every second location in each dimension. The various parameter settings are by an large those from the implementation of [Lazebnik]. Rather then calculating one sift vector at a given location and then moving onto the second location, we notice that some calculations that are performed for one location are useful for other locations.  Therefore we use the following pipeline instead.

{\bf The $x$ and $y $ derivatives.} We convolve the image with two $5\times 5$ filters that are the $x$ and $y$ derivatives of Gaussian. This results in the values of $x$ and $y$ derivatives $I_y=dI/dy, I_x=dI/dx$ of the image intensity at every location of the image. 
%The specific values of the $x$-derivative filter are the following ($y$-derivative filter is the rotation of the $x$-derivative filter): $[[0.0284,0.0261,0.,-0.0261,-0.0284],[0.1274,0.1168,0.,-0.1168,-0.1274],[0.2100,0.1927,0.,-0.1927,-0.2100] , [0.1274,0.1168,0.,-0.1168,-0.1274] , [0.0284,0.0261,0.,-0.0261,-0.0284]]$.

{\bf Orientation histogram.} This operation takes the two gradient values $I_y, I_x$ at every location and smoothly bins them into histogram of eight orientations ($0,\pi/4,\ldots,7\pi/4$) as follows. First we calculate the orientation angle $\phi=\arctan(I_y/I_x)+\pi (1-\mbox{sign}(x))/2$ and magnitude $m=\sqrt{I_x^2+I_y^2}$. Let $\phi_h(n)=n\pi/4$, $n=0,\ldots,7$. The final set of values  is $v(n)=m*\cos (\phi-\phi_h(n))_+^9$ where the $x_+=x$ if $x>0$ and $0$ otherwise. Most of these operations are computationally expensive and therefore we precompute these values. We bin the $I_y$ and $I_x$ values into $500$ bins each and for every combination ($500^2$ values) we calculate $v(n)$, $n=0,\ldots,7$. The bin range is chosen so that the values of $I_y$ and $I_x$ never fall outside the range of the binning so no checks are needed. After this computation we obtain $8$ values at every location of the image.

{\bf Smooth subsampling} We subsample the resulting features by two in each direction. Specifically let $v_{n,y,x}$ be the input value obtained from the previous step, where $n$ is the feature number and $y,x$ is the location. The output value will be $u_{n,y,x}=v_{n,2y,2x}+v_{n,2y,2x+1}+v_{n,2y+1,2x}+v_{n,2y+1,2x+1}$. This is efficient since it only involves additions. Note that it results in output values that are essentially four times larger the input values at each location. 

{\bf Smoothing} We convolve each feature with $[[1,1],[1,1]]$ filter. This is calculated using $u_{n,y,x}=v_{n,y,x}+v_{n,y,x+1}+v_{n,y+1,x}+v_{n,y+1,x+1}$ again resulting in essentially four times larger output values then input values.

{\bf Combining and normalizing into sift vector} Now we obtain $128$ component sift vector from every location of the features maps from the previous step. At every location $(x,y)$ (of the subsampled feature image) we first obtain $128$ component vector by concatenating the $8$-component vectors at the following locations $(x+2i, y+2j)$, $i=1,2,3,4$ and $j=1,2,3,4$. Then we normalize this vector as follows. If the norm of the vector is smaller then the threshold $t_h=1$ we keep the vector. If it is larger we normalize it to have size $t_h$. The result is placed into the appropriate location of the final $m_y \times m_x \times 128$ vector, where $m_{y,x} \approx n_{y,x}/2$ where $n_{x,y}$ are the dimensions of the original image. The dimensions are slightly smaller due to boundary effects. This last operation (combining and normalizing) is the most expensive operation in the sift grid calculation and we took care to implement it efficiently.
%{\bf Comparison to standard sift} The above procedure contains summing over $4\times 4$ neighborhood (the smooth subsampling followed by smoothing). 
Note that in Lazebnik's (and Lowe's original) sift the smoothing is done over a larger neighborhood with inputs near the center weighted more then those further. This makes the output more smoothly varying under translations; in our case we used equal weighting over small neighborhoods for computational efficiency. 

\subsubsection{Hashed sparse coding.} 
We used the main procedure of this paper to calculate feature vector for each sift vector. Each such computation consisted essentially of depth=$16$ multiplications of sift and tree decision vectors ($16 \times 128$ computations) followed by multiplication of the sift vector by the appropriate pseudo-inverse matrix (typically $128 \times 5$ multiplications) resulting in total of approximately $128 \times 21$ multiplications. For $2048$ dimensional feature vector this compares to $128 \times (2048+4)$ multiplications that are needed for omp resulting in almost $100$-fold reduction. Our model was trained on $2 \times 10^6$ randomly selected sift vectors from Pascal 2011 dataset. 

\subsubsection{Spatial pyramidal pooling.} 
We used the same spatial pyramidal max pooling as in [Y-Lan]. Since the feature vectors are in the sparse format the resulting computation is very efficient and negligible compared to either sift or tree sparse coding. The details are as follows. We need to calculate the maximum over the features in $1\times 1$, $2 \times 2$ and $4 \times 4$ regions of the feature vector obtained in the previous step. First we split this vector into $4 \times 4$ regions $R_{x',y'}$. Let $n_f$ be number of features, typically $2048$, $v_{f,x,y}$ be the input feature vector and $u_{f,x',y'}$, $x'=1,2,3,4$, $y'=1,2,3,4$ be the $4 \times 4$ part of the final feature vector. We calculate $u$ using the following.
\begin{equation}
u_{f,x',y'}=\mbox{max}_{x,y \in R_{x',y'}} v_{f,x,y}
\end{equation}
This calculation is done by looping over all feature vectors and indices and filling the pooled feature vector so the number of computations is of the order of the total number of nonzero features. We can get $2 \times 2$ and $1 \times 1$ parts of the final feature vector analogously. However it is more efficient now to use the $4 \times 4$ vector obtained and pool it into $2 \times 2$ regions and then pool the result into $1 \times 1$ regions. The final output vector is concatenation of these vectors, resulting in $n_f \times 21$ vector. 

\subsubsection{Classification.} Subsequently a logistic regression classifier is trained on the feature vectors using the liblinear package \cite{REF08a}. 

\subsubsection{Implementation.}
Each the following operations we implemented using a multicore processing: all steps of the sift, finding the group using tree, and multiplying by pseudo-inverses. In each of these steps separately the image/feature image was split in $n_{\mbox{cores}}$ parts and send to different core. The system was implemented in C++. 
%The -O2 flag in compilation made significant difference. 
Blas in the Accelerate framework was used in the tree sparse coding. We report the result on a macbook pro, with
a 2.3 Ghz Intel Core i7 processor with 4 cores.  The observed speedup compared to single core was about $3$. 
%\begin{table}
%\label{t:speed}
%\begin{center}
%\begin{tabular}{|c||c|c||}
%\hline
 %& 1 core (fps) & 4 cores (fps) \\
%\hline
%\hline
%SIFT & 25 & 59 \\
%\hline
%SIFT+TreeSC+pyramid  & 7 & 22.5\\		
%\hline
%\end{tabular}
%\end{center}
%\caption{Frame-rates for different parts of the system on $321 \times 481$ pixel images (frames per second) on 2011 mac book pro laptop.}
%\end{table}

We also test the run time of just the coding, compared with coding using OMP with the SPAMS package \cite{spams}.  

\subsection{Accuracy on Caltech 101 and 15 scenes}
We test the accuracy of the standard pipeline with the hashed dictionary and with standard $l_0$ sparse coding on two object recognition benchmarks, Caltech 101 \cite{} and 15 scenes \cite{}.   As mentioned before, for all data sets, we train the hashed dictionary on $2 \times 10^6$ randomly selected sift vectors from the Pascal 2011 dataset.   Caltech 101 consists of 101 image categories and approximately 50 images per category; many classes have more training examples and we do the usual normalization of error by class size.  We use 30 training examples per class.  The 15 scenes database contains 15 categories and 4485 images,
and between 200 to 400 images per category.   We use 100 training images per class on this data set.   For each data set, we run over 10 random splits and record the mean and standard deviation of the test error.   We record the results in Tables \ref{t:caltech} and \ref{t:scenes}.  The first two columns of each table correspond to the hashed sparsed coding run with $5$ or $10$ nonzero entries on Lazebnik's sift.  The next two columns correspond to the ``real time'' system, hashed sparse coding run on our approximate sift, and the last two columns correspond to OMP, trained and coded with SPAMS\cite{spams} on Lazebnik's sift.  Each row corresponds to the number of atoms in the dictionary.   As far as we know, state of the art with single features on grayscale images on Caltech 101 with 30 training examples per category is .773, in \cite{ylan-iccv-2011}, and .898 for the 15 scenes, in \cite{not_lonely}.  Both of these methods use the same basic pipeline as this work, but with variations on the sparse coding; our method can be used in conjuction with their methods.

As has been observed by other authors, increasing the size of the dictionary only seems to increase the accuracy.   Note that for our method, the only places that the size of the dictionary affects the computational cost is in training, where we use an SOMP, and in the final classification stage.   The last component is small for these experiments, but if we wanted to use the system for detection at many locations at an image, it would start to be significant.

\subsection{Running speed.} We tested the speed of the full pipeline from image to classification.  We show results on images from the Berkeley dataset and Caltech 101.
The Berkeley images are $321 \times 481$,  The Caltech 101 images were resized so that the largest size was at most 300, with the aspect ratio fixed.
With $5$ nonzero coefficients and depth $16$ tree, we get the results  in Table \ref{t:speed2}.  The entire dataset of $9145$ images in Caltech 101 was processed in $4$ minutes and $48$ seconds with $2048$ features and in $5$ minutes and $35$ seconds with $8092$ features. This corresponds to $31.75$fps and $27.3$fps respectively.

We also test the speed of just the sparse coding\footnote{This test was done on a  quad core intel i5 running 64 bit Linux, with 4 gigs of ram; both our code and SPAMS were run as a mex file through Matlab}.  Coding 15000 sift vectors with a depth 16 tree and 5 nonzeros per $x$ takes .034 seconds with one core, and .018 with four.  In comparison, SPAMS with a dictionary of size 1024 costs .898 seconds using four cores.  This is not exactly a fair test, as SPAMS must calculate a Cholesky decomposition of the Gram matrix of the dictionary when it runs, and this could be cached; however, simply multiplying the dictionary matrix by the data vectors takes .294 seconds. As the size of the dictionary increases, this will increase, but our method will not get any slower.

\begin{table*}
\label{t:caltech}
\begin{center}
\begin{tabular}{|c||c|c||c|c||c|c|| }
%\begin{tabular*}{0.2\textwidth}{@{\extracolsep{\fill}}|c||c|c|c||c|c|c|}
\hline
 & hashed $m=5$ & hashed $m=10$ & hashed $m=5$, R.T. & hashed $m=10$  R.T. & OMP $m=5$ & OMP $m=10$ \\
\hline
\hline
$K=1024$  & .722 $\pm$ .011& .704 $\pm$ .010 & .710 $\pm$ .007& .697 $\pm$ .010& .725 $\pm$ .008 & .721 $\pm$ .010 \\
\hline
$K=2048$  & .735 $\pm$ .007& .731 $\pm$ .011& .723 $\pm$ .007& .716 $\pm$ .005 & .747 $\pm$ .008& .738 $\pm$ .008\\		
\hline
 $K=4096$ & .741 $\pm$ .011& .740 $\pm$ .006& .736 $\pm$ .005& .724 $\pm$ .004 & .754 $\pm$ .008& .757 $\pm$ .010\\
\hline
 $K=8092$ & .751 $\pm$ .009&                &   .739 $\pm$ .003  &               &       &  \\
\hline
\end{tabular}
\end{center}
\caption{Caltech accuracies and standard deviations over 10 random splits.   The first two columns of each table correspond to the hashed sparsed coding run with $5$ or $10$ nonzero entries, on Lazebnik's sift.  The next two columns correspond to the ``real time'' system, hashed sparse coding run on our approximate sift, and the last two columns correspond to OMP, trained and coded with SPAMS\cite{spams} on Lazebnik's sift.  Each row corresponds to the number of atoms in the dictionary. }
\end{table*}

%\begin{table}
%\label{t:caltech_std}
%\begin{center}
%\begin{tabular}{|c||c|c||c|c|}
%%\begin{tabular*}{0.2\textwidth}{@{\extracolsep{\fill}}|c||c|c|c||c|c|c|}
%\hline
% & hashed $m=5$ & hashed $m=10$ & OMP $m=5$ & OMP $m=10$ \\
%\hline
%\hline
%$K=1024$  & .011 & .010 & .008& .010 \\
%\hline
%$K=2048$  & .007 & .011 & .008 & .008 \\		
%\hline
% $K=4096$ & .011 & ..006 & .008 & .010 \\
%\hline
%\end{tabular}
%\end{center}
%\caption{Caltech std deviation}
%\end{table}

\begin{table*}
\label{t:scenes}
\begin{center}
\begin{tabular}{|c||c|c||c|c||c|c|| }
%\begin{tabular*}{0.2\textwidth}{@{\extracolsep{\fill}}|c||c|c|c||c|c|c|}
\hline
 & hashed $m=5$ & hashed $m=10$ & hashed $m=5$, R.T. & hashed $m=10$  R.T. & OMP $m=5$ & OMP $m=10$ \\
\hline
\hline
$K=1024$  & .792 $\pm$ .006& .789 $\pm$ .004 & .786 $\pm$ .004& .770 $\pm$ .007& .801 $\pm$ .006 & .802 $\pm$ .004 \\
\hline
$K=2048$  & .807 $\pm$ .006 & .800 $\pm$ .004& .796 $\pm$ .007& .788 $\pm$ .007 & .814 $\pm$ .006& .813 $\pm$ .006\\		
\hline
 $K=4096$ &  .810 $\pm$ .007& .810 $\pm$ .004& .807 $\pm$ .003& .804 $\pm$ .004 & .826 $\pm$ .007& .822 $\pm$ .007\\
\hline
 $K=8092$ & .811 $\pm$ .004&                &   .815 $\pm$ .004             &               &       &  \\
\hline
\end{tabular}
\end{center}
\caption{15 scenes accuracies and standard deviations over 10 random splits.   The first two columns of each table correspond to the hashed sparsed coding run with $5$ or $10$ nonzero entries on Lazebnik's sift.  The next two columns correspond to the ``real time'' system, hashed sparse coding run on our approximate sift, and the last two columns correspond to OMP, trained and coded with SPAMS\cite{spams} on Lazebnik's sift.  Each row corresponds to the number of atoms in the dictionary.}
\end{table*}

\begin{table*}
\label{t:speed2}
\begin{center}
\begin{tabular}{|c||c|c||c|c||c|c|c||}
\hline
&\multicolumn{4}{c||}{$321 \times 481$ pixel images}&\multicolumn{3}{c||}{Caltech 101 (on 4 cores)} \\
\hline
 & 1 core (s) & 4 cores (s) & 1 core (fps) & 4 cores (fps) & total time (m:s) & (fps) & performance\\
\hline
\hline
SIFT & 0.039 & 0.017 & 25 & 59 & - & - & -\\
\hline
SIFT+TreeSC+pyramid & 0.143 & 0.045 & 7 & 22.5 & - & - & - \\
\hline
full (1024) & 0.145 & 0.0465 & 6.9 & 21 & 4:01 & 38 & $.710 \pm .007$\\
\hline
full (2048) & 0.1473 & 0.050 & 6.8 & 20 & 4:45 & 32 & $.723 \pm .007 $\\
\hline
full (4096) & 0.1495 & 0.052 & 6.7 & 19 & 4:42 & 32 &$ .736 \pm .005 $\\
\hline
full (8092) & 0.155 & 0.0565 & 6.4 & 18 & 5:35 & 27 & $.739 \pm .003$\\
\hline
\end{tabular}
\end{center}
\caption{Speeds of different parts of the system and different dictionary sizes on $321 \times 481$
pixel Berkeley dataset images and Caltech 101 images. The times are for single frame in seconds. Frame
rates are the inverses and are in frames per second. The total time is the time to process the entire
Caltech 101 datasets consisting of $9145$ images (minutes:seconds). The Caltech 101 images were
pre-resized so that largest side is $300$ pixels. The last column is the recognition performance when
trained on $30$ training images per category. (The speeds vary probably due to disc access and are
faster after one or more sweeps through the dataset).}
\end{table*}

\section{Conclusion}

In this paper we presented a fast approximate sparse coding algorithm and use it to build an accurate 
real time object recognition system. Our contributions can be summarized into four parts. 1) We describe a general method for learning the groups for 
greedy structured sparse coding using a generalization of LLoyd's algorithm and SOMP.
%We gave and algorithm that places each input into a given number of groups using a tree, in time
%logarithmic in the number of groups. The tree and the groups are learned from data and upon training
%similar inputs are placed in close branches in the tree. 
2) We use this method to design a fast approximation of greedy sparse coding that uses a tree structure for inference. 
%Given thus learned groups or given a set of groups given otherwise we learn a sparse coding dictionary and assignment of which dictionary elements goes into which group. The dictionary reflects the group assignment of vice versa. 
3) We give a fast
approximate implementation of the SIFT descriptor. 4) These algorithms together allow as to build a real time
object recognition system in the framework of \cite{yang-cvpr-09}. It processes the entire Caltech 101 dataset in under 5
minutes (with images resized so that larger size is $300$ pixels). As far as we know this is the first
time that a fast implementation of this type of system has been put together with comparable accuracy.

We see many possible directions in the future both for improving the group sparse coding algorithm
and applying our system to vision. We would like to learn the hash or tree, rather than build it before the dictionary training.
We would
like to train the system on larger datasets and work on real time object detection (as opposed to classification). At this speed the
algorithm allows us to process around 2 million medium sized images ($300 \times 400$) in $24$ hours on
a single computer. The object detection should also be feasible given that the expensive part –
calculation of features at different parts of the image from which detection is calculated - is fast.

{\small
\bibliographystyle{ieee}
\bibliography{cvpr2012ssc}
%\bibliography{CVPR2012}
}

\end{document}